# Retrieval-Augmented LLMs for Evidence Localization in Clinical Trial Recruitment from Longitudinal EHR Narratives

Ziyi Chen
Department of Health Outcomes and Biomedical Informatics
University of Florida
Gainesville, FL
chenziyi@ufl.edu

Mengxian Lyu
Department of Health Outcomes and Biomedical Informatics
University of Florida
Gainesville, FL
lvmengxian@ufl.edu

Cheng Peng
Department of Health Outcomes and Biomedical Informatics
University of Florida
Gainesville, FL
c.peng@ufl.edu

Yonghui Wu *
Department of Health Outcomes and Biomedical Informatics
University of Florida
Gainesville, FL
yonghui.wu@ufl.edu

## ABSTRACT

Screening patients for enrollment is a well-known labor-intensive bottleneck, causing under-enrollment and eventually trial failures. Recent breakthroughs in large language models (LLMs) offer a promising opportunity to use artificial intelligence to improve screening. This study systematically explored both encoder-based and decoder-based generative LLMs for screening clinical narratives, facilitating clinical trial recruitment. We examined both general-purpose LLMs and medical-adapted LLMs and explored three strategies to alleviate the "Lost in the Middle" issue when handling long documents, including 1) Original long-context: using the default context windows of LLMs, 2) Named Entity Recognition (NER) based extractive summarization: converting the long document into summarizations using named entity recognition, 3) Retrieval-Augmented Generation (RAG): dynamic evidence retrieval based on eligibility criteria. The 2018 n2c2 Track 1 benchmark dataset is used for evaluation. Our experimental results show that the MedGemma model with the RAG strategy achieved the best micro-F1 score of 89.05%, outperforming other models. Generative LLMs have remarkably improved trial criteria that require long-term reasoning across long documents; whereas trial criteria that span a short piece of context (e.g., lab tests) demonstrate incremental improvements. The real-world adoption of LLMs for trial recruitment has to consider specific criteria for selecting between rule-based queries, encoder-based LLMs, and generative LLMs to maximize efficiency within a reasonable computing cost.

## 1 INTRODUCTION

Clinical trials are a critical and expensive component of evidence-based medicine. A significant amount of effort is spent on the recruitment of subjects who meet the eligibility criteria[1,2]; under-enrollment is a well-known cause of trial failures. Electronic health records (EHRs) could be a valuable resource for screening subjects who meet the recruitment criteria; however, reviewing large-scale EHRs is time-consuming. Patient screening for trial recruitment has traditionally relied on manual procedures, and studies have reported that clinicians may overlook up to 60% of eligible patients[3]. More and more trials are starting to use criteria that involve information typically documented in clinical narratives, which cannot be solved with simple queries or pattern matching. Recently, large language models (LLMs) have greatly improved the use of clinical narratives for various healthcare applications. Previous studies have applied LLMs to match patients to clinical trials, using patient summarization to query a trial database and select trials based on inclusion/exclusion criteria.

This study aims to examine LLMs, RAG, and content-filtering methods to improve patient recruitment through clinical narratives. We compared three strategies to leverage long context of clinical narratives: (1) Direct long-context modeling using LLMs with natively extended context windows; (2) NER-based extractive summarization, which semantically condenses patient records by retaining key clinical entities and associated context; and (3) RAG, which dynamically retrieves relevant evidence conditioned on each eligibility criterion. This framework enables a systematic comparison of full-context processing, targeted information selection, and retrieval-based reasoning, offering insights into how LLMs can handle the long context of clinical narratives for patient recruitment.

**Contributions:** Our work's key contributions are summarized below:

• **Comparison of complementary context-retention strategies**: We propose and empirically compare three approaches for handling longitudinal clinical records for patient eligibility: direct long-context modeling, NER-based semantic condensation, and retrieval-augmented generation, highlighting their respective strengths and trade-offs.

• **Benchmarking on a real-world clinical trial dataset:** We evaluate our methods on the established N2C2 2018 shared task dataset, enabling direct comparison with prior rule-based and transformer-based systems.

• **Insights into evidence localization and reasoning in longitudinal records:** Using quantitative results and qualitative analysis, we demonstrate how different strategies capture dispersed eligibility evidence over time, providing practical guidance for deploying LLMs in real-world clinical trial recruitment workflows.

## 2 RELATED WORK

With the rapid adoption of electronic health records (EHRs) across healthcare institutions worldwide, vast amounts of patient data are now digitally stored[4,5]. EHRs contain longitudinal, rich patient information that can be leveraged for medical research[6]. Therefore, EHRs have been used to identify and pre-screen patients who are eligible for disease screening or for enrollment in clinical studies[7]. Despite the potential, using EHRs for patient recruitment encounters substantial challenges. Traditional screening approaches have relied heavily on structured data (ICD codes, demographics) and SQL-based querying[8]. However, structured data often lacks the granularity required for complex trial criteria, as an increasing number of trials require information captured in clinical narratives spanning multiple encounters. Critical eligibility-related evidence may be dispersed across various documents and time points, often requiring complex temporal reasoning and inference[9].

Natural language processing (NLP) provides a viable solution for automating eligibility screening, with the potential to improve efficiency and accelerate clinical trial recruitment. The 2018 United States national NLP clinical challenges (N2C2) shared task[10] established a benchmark for clinical trial recruitment using clinical narratives. In this national challenge, the records of 288 diabetic patients were annotated to determine whether they met 13 specific trial criteria. These criteria were chosen from real trials to require diverse NLP capabilities. Early approaches to this task were dominated by rule-based systems that combined regular expressions, heuristic textual markers, and hybrid machine learning strategies. The top-performing system achieved a micro-F1 score of 0.91[11], demonstrating the effectiveness of expert-crafted rules in capturing domain-specific patterns. These methods relied heavily on manual feature engineering and extensive clinical knowledge, limiting scalability, contextual understanding, and generalizability across institutions and trial designs.

Later, transformer-based NLP models, such as BERT[12], have transformed clinical NLP by enabling contextualized representations. Domain-specific variants, such as GatorTron[13], have consistently outperformed conventional baselines across tasks such as Named Entity Recognition (NER) and relation extraction[14,15]. Generative LLMs, such as GPT, demonstrated remarkable few-shot and zero-shot capabilities in biomedical reasoning[16,17], indicating the potential for trial recruitment. Recent studies have explored LLMs such as GPT-4 for clinical trial eligibility matching[18]. For instance, TrialGPT demonstrated that LLMs could match patients to trials with high accuracy using zero-shot prompting[19] , where a short patient profile is used to query a trial database to match the patient to eligible trials. However, this approach encountered several barriers when using EHRs for trial recruitment, as each patient typically has multiple lengthy clinical documents numbering dozens or hundreds. Although newer LLMs like Gemini 3 now support extensive context windows of over 1M tokens, the critical challenge is to identify the "useful information" related to the eligibility criteria buried in hundreds of clinical notes, let alone the cost and latency of processing hundreds of millions of clinical notes for every patient. For example, recent studies reported the "Lost in the Middle" issue[20], indicating that LLMs struggle to focus on the "useful information" when the context is exceptionally long. On the other hand, new strategies such as retrieval-augmented generation (RAG) have been developed to incorporate external knowledge into LLM outputs. In healthcare, RAG has been used to improve the factual consistency and currency of generative models by grounding them on retrieved clinical evidence. For example, a perspective by Yang et al.[21] notes that RAG "enables models to generate more reliable content" by retrieving relevant knowledge. Recent studies have demonstrated the benefit of RAG for EHRs summarization and medical question-answering, especially in reducing hallucinations[22].

## 3 PRELIMINARIES

### 3.1 Problem Definition: Clinical Eligibility Determination

The task of patient eligibility determination is formulated as a multi-label classification problem. Given a set of clinical eligibility criteria $\mathcal{C} = \{c_1, c_2, \ldots, c_n\}$ and a collection of longitudinal clinical narratives $\mathcal{D}_p$ for a patient $p$, the goal is to learn a predictive mapping $f: (\mathcal{D}_p, c_i) \rightarrow y_{p,i}$. Here, $y_{p,i} \in \{0,1\}$ represents the label, where $y = 1$ indicates the patient "meets" (Met) the criterion, and $y = 0$ indicates they do "not meet" (Not Met) the criterion. In this study, we use the 2018 N2C2 Track 1 dataset, in which $n = 13$ distinct criteria must be evaluated for each patient based on their longitudinal medical history.

## 3.2 Clinical Narratives

Clinical narrative data $\mathcal{D}_p$ consists of a temporal sequence of clinical notes $\{n_{p,t_1}, n_{p,t_2}, \ldots, n_{p,t_k}\}$ where each note $n$ contains unstructured text of varying lengths. A well-known challenge in this domain is that the total token count $\mathcal{D}_p$ may exceed the maximum context window $L_{max}$ of some Transformer-based models. For instance, while typical encoder models like BERT are restricted to $L_{max} = 512$, the average patient record in the N2C2 corpus contains approximately 5290 tokens, with some records far exceeding this mean.

## 3.3 Model Adaptation Strategies

We explored two fine-tuning strategies based on the model's parameter scale and architecture.

*3.3.1 Full Fine-tuning.* For encoder-based smaller models such as BERT and GatorTron-2k, we apply full fine-tuning to maximize their discriminative capacity. Given the relatively small parameter footprint of these architectures, ranging from 512 to 2048 tokens in context length, it is computationally feasible to update the entire set of model parameters $\Theta$. During this stage, we optimize the model by minimizing the cross-entropy loss $\mathcal{L}$ over the labeled clinical dataset $\mathcal{D}$:

$$\min_{\Theta} \sum_{(x,y)\in\mathcal{D}} \mathcal{L}(f(x;\Theta), y)$$

where $f(x;\Theta)$ represents the model's prediction for a patient record $x$ and the corresponding eligibility label $y$. This ensures that the dense representations produced by the encoder models are fully specialized to clinical note semantics.

*3.3.2 Parameter-Efficient Fine-Tuning (PEFT).* Large-size LLMs, especially decoder-only architectures such as GPT, have shown remarkable reasoning abilities for complex clinical tasks. To adapt these models to the clinical domain while maintaining computational feasibility, we utilize PEFT. Specifically, we employ Low-Rank Adaptation (LoRA)[23], which represents the weight update $\Delta W$ as a product of two low-rank matrices $A$ and $B$, such that: $W_{updated} = W + \Delta W = W + BA$, where $B \in R^{d\times r}$ and $A \in R^{r\times k}$ with rank $r \ll \min(d,k)$. This allows the model to learn task-specific eligibility logic without the memory overhead of updating billions of parameters.

# 4 METHODOLOGY

We compared three strategies: 1) Original Long-Context: Using the default context windows of LLMs, 2) NER-Based extractive summarization: converting the long-document into summarizations using named entity recognition, and 3) RAG: Dynamic evidence retrieval based on eligibility criteria. This design enables systematic comparison between heuristic truncation, targeted information selection and full-context modeling.

## 4.1 NER-Based Extractive Summarization

To reduce input length while preserving clinically salient information, we implement an NER-driven note condensation strategy, a core NLP task that identifies predefined semantic categories in text. We employ a clinical NER model based on the GatorTron model fine-tuned on the i2b2 dataset that recognizes key medical entities, including problems, treatments, and tests. Per the NER extraction results, we extract sentences containing at least one problem entity, as these are most informative for eligibility assessment. The selected sentences are concatenated to form a shortened summary. If the resulting text exceeds the model's token limit, we truncate the sequence to the maximum allowable length. This approach simulates a realistic scenario in which more recent clinical notes are prioritized, and earlier history may be omitted from very lengthy records.

## 4.2 Retrieval-Augmented Generation (RAG)

We explored RAG to identify "useful" content that is related to the trail criteria. We implement an RAG pipeline that explicitly guides the model toward criterion-relevant evidence. The key intuition is that, for a given eligibility criterion, only a small subset of a patient's longitudinal notes is typically relevant, and exposing the LLM to these targeted excerpts can improve both accuracy and interpretability.

Our RAG pipeline consists of three stages. First, all patient notes in the training corpus are segmented into semantically coherent chunks. Next, each chunk is encoded into a dense vector using the open-source embedding models 'BAAI/bge-large-en-v1.5' and 'BAAI/bge-small-en-v1.5'[24], which are particularly effective for semantic search, document retrieval, and similarity comparison. This approach balances representation quality with computational efficiency. Finally, the resulting embeddings are stored in a FAISS vector index[25], while maintaining the mappings between chunks and patients.

During the retriever stage, given a patient's records $\mathcal{D}_p$ and a specific eligibility criterion $C_i$, we construct a query $Q$ using the criterion text along with the patient $P$ identifier. The query is encoded with the same embedding model, and cosine similarity is used to retrieve the top-$k$ most relevant note chunk embeddings $E(d_{p,j})$ for that patient. The retrieved snippets are assembled into a structured prompt together with the criterion description.

## 4.3 Fine-tuning Stage

The fine-tuning stage aims to instruct LLMs to determine if a subject "met" or "not met" the eligibility criteria of a trial. We evaluate both encoder- and decoder-based LLMs for this task. For the encoder-based models, we use BERT and GatorTron-2k, a 2,048-token variant of the GatorTron architecture pre-trained on a large-scale corpus of University of Florida Health clinical notes. For the decoder-based models, our evaluation includes LLaMA 3.1 8B, which offers enhanced reasoning and an extended context window (8k tokens in our configuration); GPT-OSS[26], a 20B-parameter open-weight model optimized for scalable fine-tuning; and MedGemma 27B[27], a domain-specific model tailored for medical applications.

To ensure computational feasibility when adapting these LLMs, we use LoRA, which injects trainable low-rank matrices into the attention and projection layers, enabling task-specific adaptation without the prohibitive cost of updating full model weights. This approach significantly reduces GPU memory overhead and training time while maintaining competitive performance. To maximize model capacity, we fine-tune encoder-based models with full fine-tuning. We apply FlashAttention[28] across all experiments to optimize memory efficiency and computational throughput.

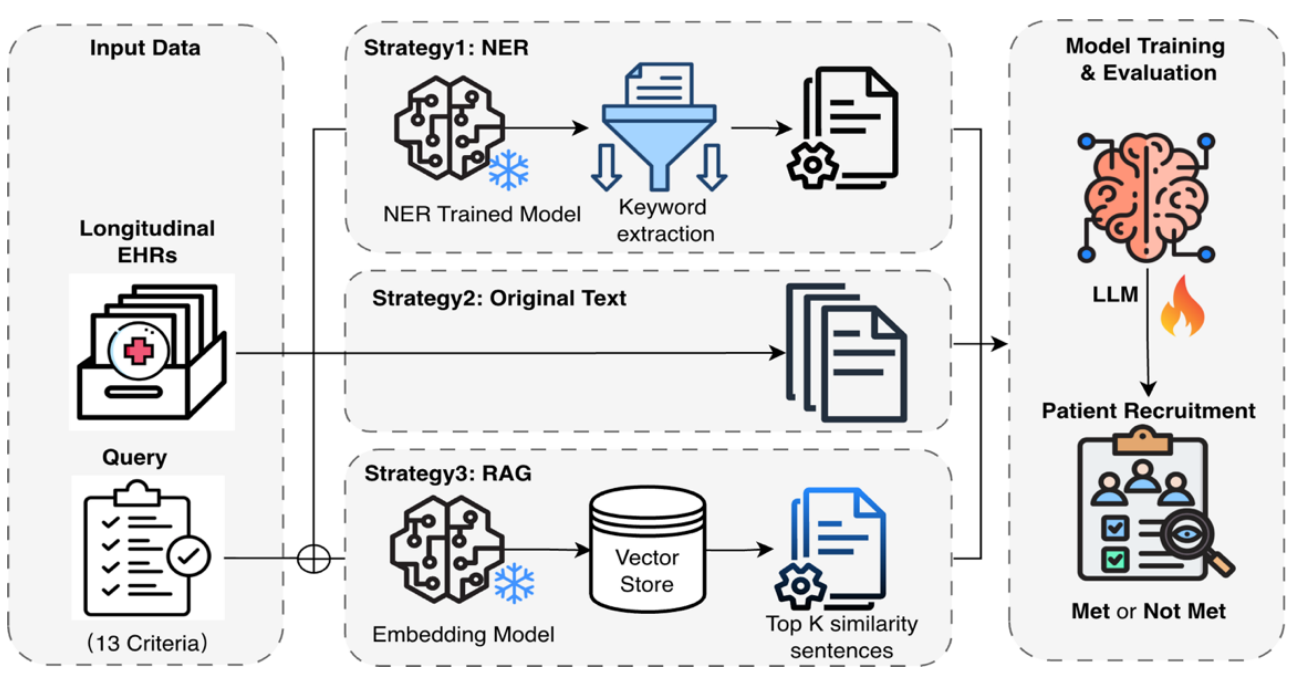

**Figure 1. Framework of LLM-driven patient recruitment strategies using longitudinal EHRs.**

# 5 EXPERIMENTS

## 5.1 Experimental Settings

*5.1.1 Dataset Description.* We use the 2018 N2C2 Track 1 dataset, focusing on Clinical Trial Cohort Selection. The dataset consists of medical records for 288 unique diabetic patients derived from the 2014 i2b2/UTHealth corpus. These patients are all diabetics at risk of cardiovascular diseases, and the criteria reflect common trial conditions related to diabetes and cardiovascular health. The 13 eligibility criteria were derived from real clinical trials. Medical professionals manually reviewed the documents to assign a binary label indicating whether the subject met the eligibility criteria. In total, there are 13 criteria, including examples such as ADVANCED-CAD (advanced cardiovascular disease based on multiple sub-conditions), ALCOHOL-ABUSE (excessive alcohol use), MAJOR-DIABETES (major diabetes-related complications), and MI-6MOS (myocardial infarction in the last 6 months).

Each patient has 2–5 notes, averaging ~2,711 tokens per patient (note lengths vary widely, from a few hundred to several thousand tokens). We use the official Train and Test split, then randomly split the validation dataset from the training set: 161 patients for training, 41 for validation, and 86 for testing. There is a class imbalance across criteria: some conditions are prevalent, while others are rare. **Table 1** shows the distribution of Met/Not Met counts per criterion across the 288 patients. We use micro-averaged F1 to evaluate the overall performance.

*5.1.2 Implementation Details.* We implemented all models using PyTorch and the HuggingFace Transformers library. We employed LoRA with a rank r=8 and alpha=64. The optimization process uses the AdamW optimizer with a learning rate of 1e-4 and a batch size of 4, and incorporates early stopping based on validation performance to prevent overfitting. For the RAG pipeline, we set the retrieval parameter k=10, which we empirically found consistently captures clinically significant evidence. Model inference is guided by a unified instruction-based prompting strategy that concatenates structured EHR narratives and specific queries into a single context window; query structures and formal templates are detailed in **Appendix Tables 1 and 2**. All experiments were conducted using two NVIDIA B200 GPUs, each with 192GB of memory.

*5.1.3 Evaluation metrics.* We use the official evaluation scripts released by the N2C2 2018 Track 1 to calculate the evaluation scores. We report the Area Under the Receiver Operating Characteristic Curve (AUROC) and micro-averaged F1-score as our primary indicators of discriminative capability and overall generation accuracy. To further capture the impact of class imbalance across the 13 eligibility criteria, we include macro-averaged F1-scores, micro-precision, and micro-recall.

## 5.2 Experiment Results

*5.2.1 Data Characteristics and Processing.* **Table 1** shows the 288 subjects included in the 2018 N2C2 dataset, partitioned into training (n=161), validation (n=41), and test (n=86) sets, stratified by 13 distinct trial criteria. The distribution varies significantly. For instance, MAKES-DECISIONS and ENGLISH show high prevalence of 96.2% and 92.0%, respectively. In comparison, KETO-1YR represents an extreme minority at only 0.3% (n=1), ALCOHOL-ABUSE at 3.5%, and DRUG-ABUSE at 5.2%, indicating an extremely imbalanced dataset. **Table 2** presents the token distribution, highlighting the computational challenges of the original text (mean 5,290; max 15,186). The NER strategy reduces the mean number of tokens to 2,955. The RAG-top10 strategy further reduced the mean number of token count to 1,403.

**Table 1. Met /Not Met (Y/N) counts per trial criterion across the 288 patients**

| **Criterions** | **Total (n=288)** | |
|---|---|---|
| | Y (%) | N (%) |
| ABDOMINAL | 107 (37.2%) | 181 (62.8%) |
| ADVANCED-CAD | 170 (59.0%) | 118 (41.0%) |
| ALCOHOL-ABUSE | 10 (3.5%) | 278 (96.5%) |
| ASP-FOR-MI | 230 (79.9%) | 58 (20.1%) |
| CREATININE | 106 (36.8%) | 182 (63.2%) |
| DIETSUPP-2MOS | 149 (51.7%) | 139 (48.3%) |
| DRUG-ABUSE | 15 (5.2%) | 273 (94.8%) |
| ENGLISH | 265 (92.0%) | 23 (8.0%) |
| HBA1C | 102 (35.4%) | 186 (64.6%) |
| KETO-1YR | 1 (0.3%) | 287 (99.7%) |
| MAJOR-DIABETES | 156 (54.2%) | 132 (45.8%) |
| MAKES-DECISIONS | 277 (96.2%) | 11 (3.8%) |
| MI-6MOS | 26 (9.0%) | 262 (91.0%) |

**Table 2. Data token size information**

| Input Type | Mean Tokens [Min, Max] | Applied Token Limit |
|---|---|---|
| NER-problem | 2955 [235, 10120] | 512/ 2048/ 8192 |
| Ori Text | 5290 [1670, 15186] | 8192 |
| RAG-top10 | 1403 [777, 2777] | 2048 |

*5.2.2 Performance Comparison Across Modeling Strategies.* **Table 3** compares the performance of the three strategies. The RAG approach generally yielded the most robust results, particularly when using the BAAI-large embedding model, which consistently showed marginal gains over the BAAI-small variant across all metrics for MedGemma-27b.

Interestingly, the full-context strategy without RAG is highly competitive, achieving a micro-F1 score of 0.8849 for MedGemma-27b. The RAG strategy's ability to selectively retrieve relevant evidence provided a clearer signal to the generator, ultimately yielding the highest observed scores, as shown in **Figure 2**. In contrast, the NER-based filtering strategy generally underperformed compared to the other two methods. For the Llama 3.1 8b model, the micro-F1 score decreased from 0.8671 in the full-context setting to 0.8293 with NER filtering, suggesting that the NER procedure missed important information compared with RAG.

As for the models, MedGemma-27b consistently achieves the best performance across all strategies, achieving the best micro-F1 of 0.8905 and an AUC of 0.8922 under the RAG [BAAI-large] configuration. In contrast, encoder-based models such as BERT and GatorTron-2k showed lower performance, with BERT reaching only 0.7240 in micro-F1 and 0.7260 in AUC. Our results demonstrate the advances of decoder-based LLMs.

*5.2.3 Per-criterion F1 Performance.* **Table 4** shows detailed performance for each trial criterion. The experimental results show that RAG-based generative LLMs benefit more for trial criteria that require long-distance, across multi-document reasoning, such as ALCOHOL-ABUSE, DRUG-ABUSE, and ENGLISH, with absolute F1 improvements exceeding 40% compared to the best NER-512 baseline. Trial criteria that do not need long-distance reasoning, such as clinical measurements (e.g., CREATININE, MI-6MOS), show minimal incremental gains from RAG. Criteria such as KETO-1YR remain unchanged across all strategies due to extreme class imbalance and limited documentation in clinical notes. Overall, RAG improves micro-F1 by 16.65% and macro-F1 by 30.37% relative to NER-512, demonstrating both enhanced accuracy and improved robustness across heterogeneous eligibility criteria.

**Table 3. Overall performance comparison of encoder-based and decoder-based models across NER, original long-context, and RAG strategies**

| | Finetune Strategy | Generator Models | Token Limit | Micro -F1 | Micro -Prec. | Micro -Rec. | AUC |
|---|---|---|---|---|---|---|---|
| NER | Full Finetune | BERT | 512 | 0.7240 | 0.6632 | 0.6993 | 0.7260 |
| | | GatorTron-2k | 2048 | 0.7423 | 0.6900 | 0.7081 | 0.7433 |
| | LoRA | Llama 3.1 8b-it | 8192 | 0.8293 | 0.7591 | **0.8649** | 0.8369 |
| | | GPT-oss20b | 8192 | 0.7453 | 0.6920 | 0.7146 | 0.7465 |
| | | MedGemma 27b-it | 8192 | **0.8629** | **0.8284** | 0.8519 | **0.8645** |
| ORI | LoRA | Llama 3.1 8b-it | 8192 | 0.8671 | 0.8051 | 0.8998 | 0.8740 |
| | | GPT-oss20b | 8192 | 0.7910 | 0.7468 | 0.7647 | 0.7921 |
| | | MedGemma 27b-it | 8192 | **0.8849** | **0.8337** | **0.9063** | **0.8902** |
| RAG | LoRA [BAAI-small] | Llama 3.1 8b-it | 8192 | 0.8533 | 0.7755 | **0.9107** | 0.8634 |
| | | GPT-oss 20b | 8192 | 0.8172 | 0.7720 | 0.8039 | 0.8193 |
| | | MedGemma 27b-it | 8192 | **0.8880** | **0.8515** | 0.8889 | **0.8906** |
| | LoRA [BAAI-large] | Llama 3.1 8b-it | 8192 | 0.8540 | 0.8265 | 0.8301 | 0.8542 |
| | | GPT-oss20b | 8192 | 0.8117 | 0.7657 | 0.7974 | 0.8137 |
| | | MedGemma 27b-it | 8192 | **0.8905** | **0.8602** | **0.8845** | **0.8922** |

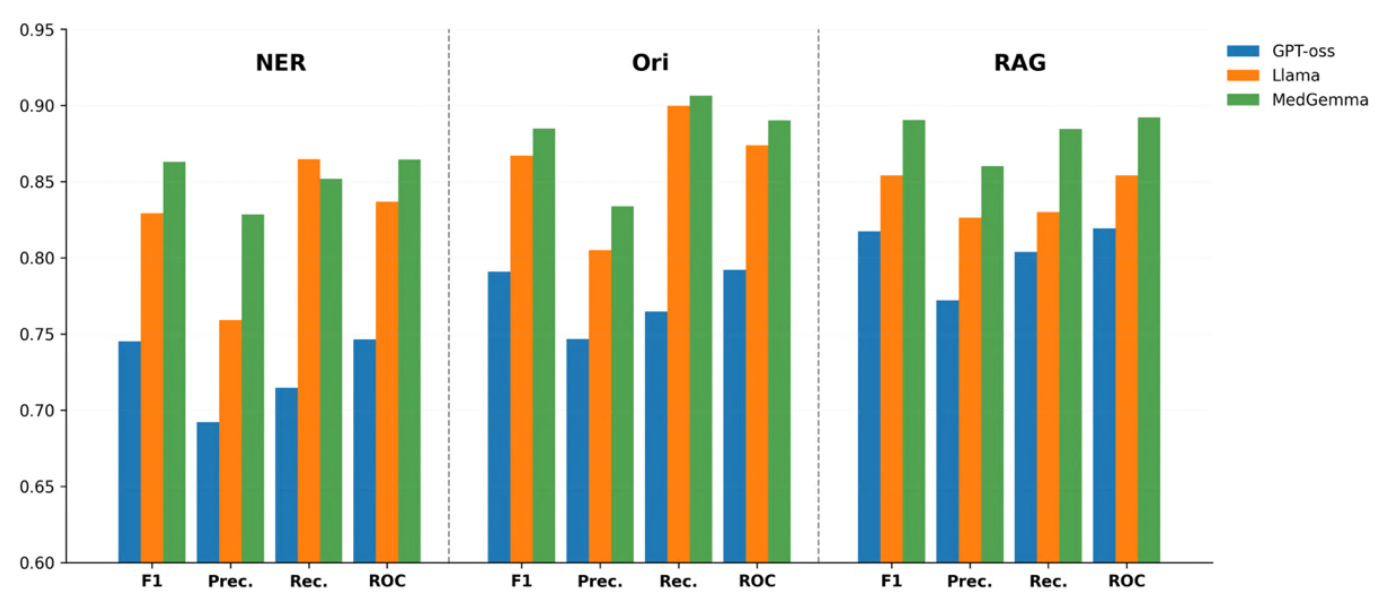

**Figure 2. Overall performance comparison of decoder-based models across NER, original long-context, and RAG strategies**

**Table 4. Per-criterion micro-F1 performance showing the differential impact of context management on specific clinical categories**

| **Criterion** | NER-512 | NER-2k | LongCtx-8k | RAG-top10 | Δ (RAG top10−NER-512) |
|---|---|---|---|---|---|
| ABDOMINAL | 0.5006 | 0.7728 | 0.7433 | 0.8032 | 0.3026 |
| ADVANCED-CAD | 0.5023 | 0.8136 | 0.7878 | 0.8227 | 0.3204 |
| ALCOHOL-ABUSE | 0.4911 | 0.6910 | 0.7440 | 0.8970 | 0.4059 |
| ASP-FOR-MI | 0.5554 | 0.5826 | 0.7734 | 0.7719 | 0.2165 |
| CREATININE | 0.3864 | 0.7574 | 0.8500 | 0.8044 | 0.418 |
| DIETSUPP-2MOS | 0.4983 | 0.7209 | 0.7628 | 0.8010 | 0.3027 |
| DRUG-ABUSE | 0.485 | 0.8689 | 0.8241 | 0.9255 | 0.4405 |
| ENGLISH | 0.4591 | 0.7737 | 0.9547 | 0.9780 | 0.5189 |
| HBA1C | 0.5083 | 0.8103 | 0.8201 | 0.8567 | 0.3484 |
| KETO-1YR | 0.5000 | 0.5000 | 0.5000 | 0.5000 | 0 |
| MAJOR-DIABETES | 0.464 | 0.8136 | 0.7906 | 0.7906 | 0.3266 |
| MAKES-DECISIONS | 0.4911 | 0.6277 | 0.6910 | 0.5896 | 0.0985 |
| MI-6MOS | 0.4756 | 0.7312 | 0.8752 | 0.7244 | 0.2488 |
| **Micro avg** | 0.7240 | 0.8629 | 0.8849 | 0.8905 | **0.1665** |
| **Macro avg** | 0.4859 | 0.7280 | 0.7782 | 0.7896 | **0.3037** |

## 6 CONCLUSION

This study examines LLMs and strategies for handling long clinical narratives in clinical trial recruitment. We systematically examined widely used encoder-based and decoder-based models from both the general and medical domains. Our experimental results demonstrate that generative LLMs such as MedGemma-27b, coupled with RAG with high-density embedding models, achieve remarkable advancement for navigating the complex long medical records for clinical trial recruitment. The best system based on MedGemma-27B reduced computational cost from an average of 5,290 tokens to 1,403 while achieving the best micro-F1 of 89.05%. Our experimental results support the adoption of RAG-based generative LLMs for clinical trial recruitment, outperforming traditional NER-based filtering.

We observed benefits from domain adaptation for both encoder-based and decoder-based LLMs. Decoder-based medical LLM, MedGemma, outperformed other general-purpose decoder-based LLMs; whereas encoder-based medical LLM, GatorTron, outperformed other general-purpose encoder models. Per the trial criteria, our results show that generative LLMs with RAG benefit more for criteria that require long-term reasoning across evidence dispersed across a lengthy document, such as social history and behavioral assessments. Trial criteria that can be determined in a short context, such as lab tests, usually yield incremental improvements. Despite the inherent challenges posed by imbalanced classes, the combination of LoRA-based fine-tuning and structured instruction prompting achieved robust discriminative performance.

In summary, this research demonstrates the effectiveness of RAG-enhanced generative LLMs as an efficient tool for automating the labor-intensive process of screening patients for clinical trial recruitment. This study has limitations. To facilitate replication, we used the 2018 N2C2 dataset, available through a data use agreement, which is a good approximation of real-world EHRs but still cannot reflect the full complexity. We will explore EHRs in our healthcare system at UF Health to further examine the proposed models. Future work should also explore advanced solutions based on agentic LLM models.

## ACKNOWLEDGMENTS

This study was partially supported by grants from the Patient-Centered Outcomes Research Institute® (PCORI®) Award (ME-2018C3-14754, ME-2023C3-35934), the PARADIGM program awarded by the Advanced Research Projects Agency for Health (ARPAH), National Institute on Aging, NIA R56AG069880, National Institute of Allergy and Infectious Diseases, NIAID R01AI172875, National Heart, Lung, and Blood Institute, R01HL169277, National Institute on Drug Abuse, NIDA R01DA050676, R01DA057886, the National Cancer Institute, NCI R37CA272473, and the UF Clinical and Translational Science Institute. The content is solely the responsibility of the authors and does not necessarily represent the official views of the funding institutions. We gratefully acknowledge the support of NVIDIA Corporation and the NVIDIA AI Technology Center (NVAITC) UF program.

## A APPENDIX

| **Appendix Table 1. Criteria descriptions** |
|---|
| **DRUG-ABUSE**: Drug abuse, current or past |
| **ALCOHOL-ABUSE**: Current alcohol use over weekly recommended limits |
| **ENGLISH**: Patient must speak English |
| **MAKES-DECISIONS**: Patient must make their own medical decisions |
| **ABDOMINAL**: History of intra abdominal surgery, small or large intestine resection or small bowel obstruction |
| **MAJOR-DIABETES**: Major diabetes-related complication as any of the following that are a result of (or strongly correlated with) uncontrolled diabetes: Amputation, Kidney damage, Skin conditions, Retinopathy, Nephropathy, Neuropathy |
| **ADVANCED-CAD**: Advanced cardiovascular disease having two or more of the following: Taking two or more medications to treat CAD, History of myocardial infarction, Currently experiencing angina, Ischemia, past or present |
| **MI-6MOS**: Myocardial infarction in the past 6 months |
| **KETO-1YR**: Diagnosis of ketoacidosis in the past year |
| **DIETSUPP-2MOS**: Taken a dietary supplement excluding Vitamin D in the past 2 months |
| **ASP-FOR-MI**: Use of aspirin to prevent myocardial infarction |
| **HBA1C**: Any HbA1c value between 6.5% and 9.5% |
| **CREATININE**: Serum creatinine > upper limit of normal |

| **Appendix Table 2. Prompt template** |
|---|
| Context:<br>[Longitudinal EHRs]<br><br>Question:<br>Based on the patient's medical records provided, assess if the patient meets the criteria for ……<br>Only respond with 'met' if the criteria are met or 'not met' if they are not. |